\documentclass[letterpaper, 11 pt, peerreviewca]{ieeeconf}
\usepackage{courier}
\usepackage{multicol}
\usepackage{hyperref}
\usepackage{graphicx,subfigure}
\usepackage{epstopdf}
\usepackage{tikz}
\usepackage{amsmath}
\usepackage{multirow}
\usepackage{amsfonts}
\usepackage{float}
\usepackage{amssymb}
\usepackage{graphicx}
\usepackage{flushend}
\usepackage[ruled]{algorithm}
\usepackage{algpseudocode}
\usepackage{nccmath}
\usepackage{caption}
\usepackage{lipsum}
\usepackage[rightcaption]{sidecap}

\newtheorem{theorem}{Theorem}

\newtheorem{proposition}{Proposition}

\newtheorem{definition}{Definition}
\newtheorem{remark}{Remark}

\makeatletter

\newcommand{\Rmnum}[1]{\expandafter\@slowromancap\romannumeral #1@}

\makeatother

\begin{document}
%\begin{frontmatter}
%\parbox[0.1in]{1.1in}{\raggedright Because the graph is a line on this semilog paper, the relationship is exponential.}
\title{\LARGE Visual Navigation Using Sparse Optical Flow and {\it Time-to-Transit}}
\author{Chiara Boretti, Philippe Bich, Yanyu Zhang, and John Baillieul}
\maketitle
\let\thefootnote\relax\footnotetext{\noindent
\hspace{-0.1in}\hrulefill \\
C.B.\ and P.B.\ are with Politecnico di Torino, Department of Electronics and Telecommunications (DET).  Y.Z.\ and J.B.\ are with the College of Engineering at Boston University.  \\ Support from various sources including the U.S.\ Office of Naval Research grants N00014-10-1-0952, N00014-17-1-2075, and N00014-19-1-2571 is gratefully acknowledged.  The authors are also grateful for support from Politecnico di Torino,  DET. The authors would like to thank Roy Xing for his precious help with the robot experiments at Boston University.}
%A condensed version of this paper has been submitted to the xxx} 

\begin{abstract}
\noindent Drawing inspiration from biology, we describe the way in which visual sensing with a monocular camera can provide a reliable signal for navigation of mobile robots.  The work takes inspiration from a classic paper by Lee and Reddish ({\em Nature}, 1981, https://doi.org/10.1038/293293a0) in which they outline a behavioral strategy pursued by diving sea birds based on a visual cue called {\em time-to-contact}.  A closely related concept of {\em time-to-transit}, $\tau$, is defined below, and it is shown that idealized steering laws based on monocular camera perceptions of $\tau$ can reliably and robustly steer a mobile vehicle within a wide variety of spaces in which features perceived to lie on walls and other objects in the environment provide adequate visual cues.  The contribution of the paper is two-fold.  It provides a simple theory of robust vision-based steering control.  It goes on to show how the theory guides the implementation of robust visual navigation using ROS-Gazebo simulations as well as deployment and experiments with a camera-equipped Jackal robot.  As far as we know, the experiments described below are the first to demonstrate visual navigation based on $\tau$.
\end{abstract}
\begin{flushleft} {{\bf Keywords}: Time-to-transit, Eulerian optical flow, Lagrangian optical flow, vision-based navigation} 
\end{flushleft}

\section{Introduction}
The terminology {\em time-to-contact} and the notation $\tau$ appear to have originated among researchers working in the Cornell University lab of James J. Gibson in the 1970's and 1980's (\cite{lee1976theory},\cite{lee1981plummeting}).  Since then, $\tau$ has been a significant focus of research in perceptual psychology (e.g.\ \cite{baures2021time},\cite{tresilian1995perceptual}). The basic idea is simple. When an observer is approaching an object at a constant velocity, the image of the object on the observer's retina (or camera's image plane) is an expanding ``bundle'' of vectors, and if we take any particular image point $r(t)$ (in image plane coordinates) and differentiate with respect to time, then as noted in \cite{lee1981plummeting}, $r/\dot r$ is equal to $\tau$---the {\em time-to-contact} or {\em time-to-collision}.  In order to explore how $\tau$ might be used as a navigation signal, we recently introduced a generalization called {\em time-to-transit} (\cite{sebestabaillieul},\cite{kong2013optical}).  To understand how time-to-transit can be perceived and used to guide navigation, we consider the geometry of planar motion of a camera-equipped unicycle vehicle relative to a feature point in the environment.  The kinematics of the vehicle are:
\begin{equation}
\left(\begin{array}{c}
\dot x \\
\dot y \\
\dot\theta\end{array}\right) = \left(\begin{array}{c}
v\cos\theta \\
v\sin\theta \\
u\end{array}\right),
\label{eq:jb:BasicVehicle}
\end{equation}
where $v$ is the forward speed in the direction of the body-frame $x$-axis, and $u$ is the turning rate.  

\begin{figure}[h]
\begin{centering}
\includegraphics[scale=0.4]{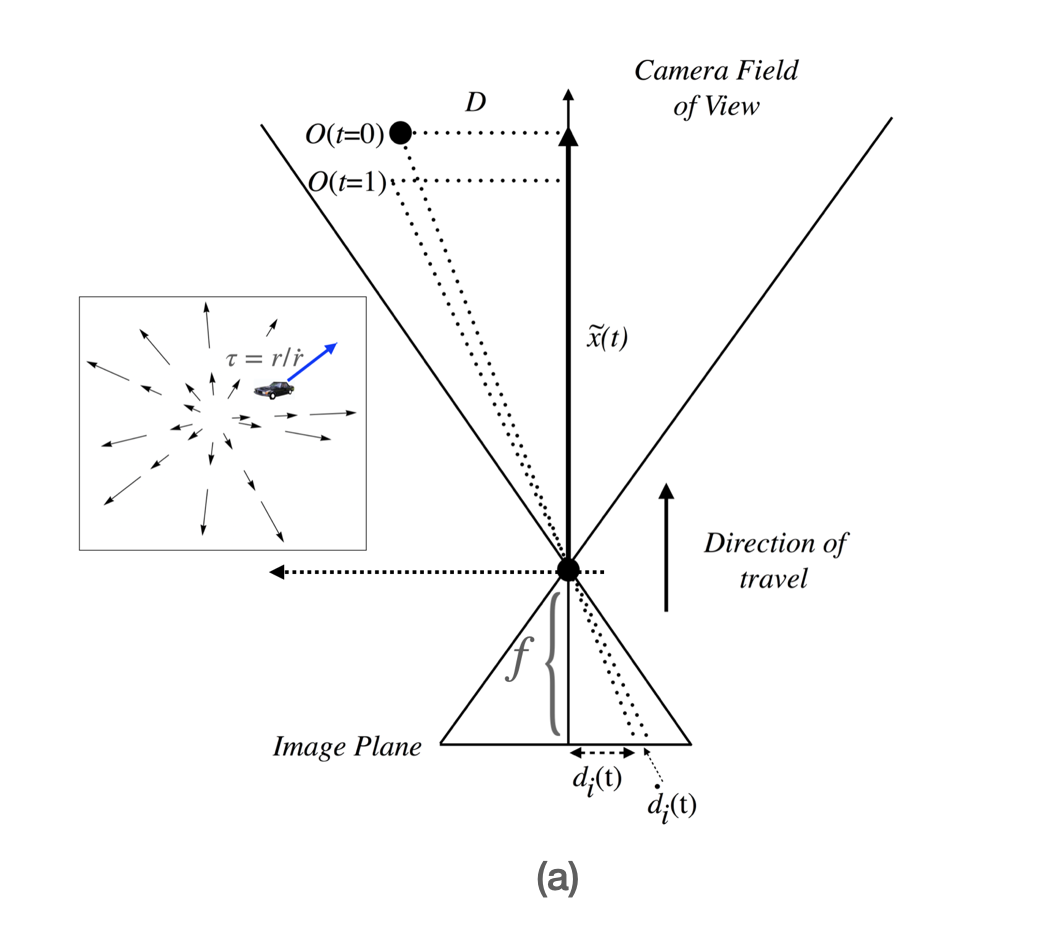}
\includegraphics[scale=0.4]{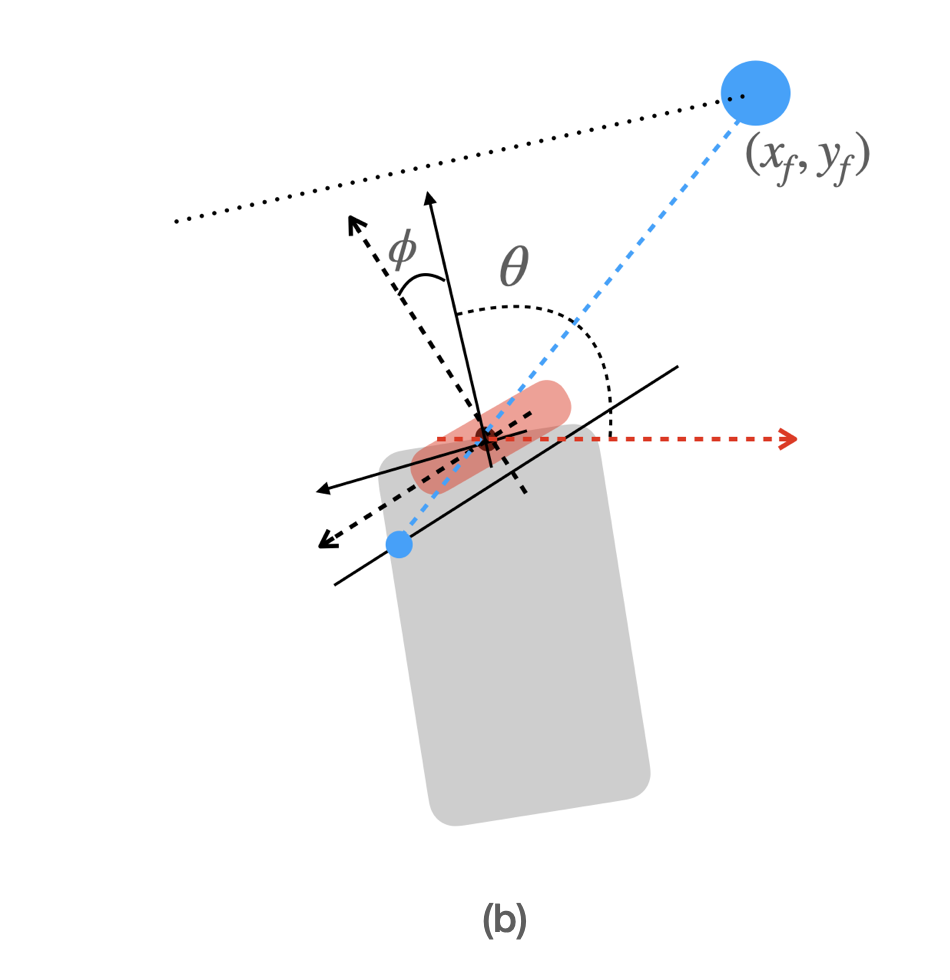}
\caption{When $\phi=0$, $\tau$ is a proxy for distance or depth. If the forward-looking camera is directly abeam of the feature point to its right or left, the value of $\tau$ is zero; it is the instant at which the transit occurs.  If the camera heading is directly toward the feature point, the value of $\tau$ is maximized relative to all other possible headings.  If $\phi\ne 0$, as in (b), then $\tau_{per}\ne\tau_{geom}$, but for $|\phi|$ small and constant, the difference in these $\tau$-values is modest. In the inset in (a), constant speed forward motion of a forward-looking camera results in an expanding bundle of velocity vectors produced by the motions of feature images.  The quantities $r/\dot r$ in the inset and $d_i/\dot d_i$ in the large figure, are the perceived {\em time-to-transit}.}
\label{fig:jb:geometricTau}
\end{centering}
\end{figure}

\begin{definition}
Consider a feature point with coordinates $(x_f,y_f)$ and a vehicle whose body-frame configuration is as depicted in Fig.\ \ref{fig:jb:geometricTau}.  Given the current configuration $(x(t),y(t),\theta(t))$  and unicycle kinematics (\ref{eq:jb:BasicVehicle}), the {\em geometric time-to-transit} ({\em geometric} $\tau$) is the time it would take the vehicle to cross a line intersecting the feature and perpendicular to the current heading under the assumption that speed $v$ and heading $\theta$ are held constant.
\begin{flushright}$\Box$
\end{flushright}
\end{definition}

%\begin{figure}[h]
%\begin{center}
%\includegraphics[scale=0.25]{PerceivedTau.png}
%%\includegraphics[scale=0.3]{PerceivedTauB.png}
%\end{center}
%\caption{In the visual field, constant speed forward motion of a %forward-looking camera results in an expanding bundle of velocity vectors %produced by the motions of feature images (top inset).  If the magnitude of %the pixel coordinate relative to the focus of expansion is divided by the rate %of change of this magnitude ($r/\dot r$ in the inset and $d_i/\dot d_i$ in the %large figure), the result is the {\em time-to-transit}.}
%\label{fig:jb:PerceivedTau}
%\end{figure}

Simple plane geometry shows that geometric $\tau$, defined in this way, is given by
\begin{equation}
\tau(t) = \frac{\cos\theta(x_f-x(t))+\sin\theta(y_f-y(t))}{v}.
\label{eq:jb:GeomTau}
\end{equation}
Time-to-transit, $\tau$, is of interest as a navigation signal because under the assumption of constant speed $v$ and heading $\theta$, this geometric value of $\tau$ can be perceived on the image plane or retina as $r/\dot r$, which in the case of a forward-looking camera, 
%if we let $\tilde x(t) = %\cos\theta(x_f-x(t))+\sin\theta(y_f-y(t))$, and assume that %$\theta$ and $v$ are constant heading and speed %respectively, Fig.\ \ref{fig:jb:PerceivedTau} illustrates a %simple geometric argument using similar triangles showing %that: \[
%\frac{D}{\tilde x(t)}=\frac{d_i(t)}{f},
%\]
%and from this it follows that: 
is just $\tau =d_i(t)/{\dot d_i(t)}$ as depicted in Fig.\ \ref{fig:jb:geometricTau}.
%\begin{equation}
%\tau = %\frac{\tilde x(t)}{v}=
%\frac{d_i(t)}{\dot d_i(t)},
%\label{eq:jb:perceived}
%\end{equation}
%which is equal to the geometric quantity of equation (\ref{eq:jb:GeomTau}).  
Because equality only holds under the assumptions that $\theta$ and $v$ are constant and the camera is forward facing with the optical axis being aligned with the direction of motion,
%For this reason, it is important to distinguish the quantity in equation (\ref{eq:jb:perceived}) from the geometric quantity in equation (\ref{eq:jb:GeomTau}), and 
we call this value {\em perceived time-to-transit} and denote it by $\tau_{per}$.  When these ideal assumptions (constant speed, heading, and camera angle) don't hold, $\tau_{per}$ is distorted relative to $\tau_{geom}$.   Referring to Fig.\ \ref{fig:jb:geometricTau}(b), with $\phi$ not necessarily equal to 0, the general form of (\ref{eq:jb:GeomTau}) is given by
\begin{equation}
\medmath{\tau^*=\left[\left(x_f-x(t)\right) \sin (\theta +\phi )-\left(y_f-y(t)\right) \cos
   (\theta +\phi )\right)] 
\cdot \frac{\left[\left(x_f-x(t)\right) \cos (\theta +\phi
   )+\left(y_f-y(t)\right) \sin (\theta +\phi )\right]}{v\,[\sin (\theta )
   \left(x_f-x(t)\right)-\cos (\theta ) \left(y_f-y(t)\right)]}.}
 \label{eq:jb:tauPer2}
\end{equation}%Details in LauraProject/Camera_Orientation_and_Tau.nb
The extent of the distortion caused by the camera's misalignment with the vehicle heading can be illustrated 
by having the vehicle (\ref{eq:jb:BasicVehicle}) follow a straight line path from the origin of the global coordinate frame $(x(0),y(0))=(0,0)$ along the positive global frame $y$-axis---$(x(t),y(t),\theta)=(0,t,\pi/2)$.  We assume that the feature point (whose global frame coordinates are $(x_f,y_f)$ (Fig.\ \ref{fig:jb:geometricTau}(b)) is located such that $x_f>0$, $y_f>0$, and that $v=1$.  Substituting these values into equation (\ref{eq:jb:tauPer2}) yields:

\begin{align}
\begin{split}
\medmath{\tau_{per}(t)} &= \medmath{t^2 \frac{ \sin (\phi ) \cos (\phi )}{x_f}} +\\
&+ \medmath{t \left(-\frac{2 y_f \sin (\phi ) \cos (\phi)}{x_f}+ \sin ^2(\phi )-\cos ^2(\phi )\right)} +\\
   &+ \medmath{\frac{y_f^2 \sin (\phi ) \cos (\phi )}{x_f}-x_f \sin (\phi ) \cos (\phi )} +\\
   &- \medmath{y_f \sin ^2(\phi )+y_f \cos ^2(\phi )}\\[0.07in]
   &= \medmath{y_f-t + \frac{\phi}{x_f}(y_f-t)^2 - \phi x_f + o(\phi)}.
 \label{eq:jb:canonicalSim}
\end{split}
\end{align}

%\begin{equation}
%    \medmath{\tau_{per}(t)}=\medmath{y_f-t + \frac{\phi}{x_f}(y_f-t)^2 - \phi x_f + o(\phi)}.
 %   \label{eq:jb:canonicalSim}
%\end{equation}

When $\phi = 0$, this value yields the linear decrease $\tau_{per}(t)=y_f-t$, which is equal to $\tau_{geom}$.  When $\phi\ne 0$, the expression for $\tau_{per}(t)$ involves a second order term, whose effect, as illustrated in (\ref{eq:jb:canonicalSim}) is small if $\phi$ has small magnitude.  The value of $\tau_{per}$ will be more severely distorted if the assumption of constant heading is relaxed, and this will be discussed in detail in the next section.

The remainder of the paper is organized as follows.  In the next section, we briefly describe how a highly idealized model of optical flow and time-to-transit can provide a robust navigation signal.  We briefly discuss factors that confound visual signals in implementations, and by considering how perceived values of $\tau$ become distorted from the ideal, we are led to mitigation strategies that make $\tau$-based steering feasible.  Implementations using resources from ROS, Gazebo, and OpenCV are discussed in Section \ref{sec:sense-act}.  Selected details of a theory of $\tau$-based motion control are presented in Section \ref{sec:motion-control}, and experiments involving both simulations and tests with a Jackal robot are presented.  The paper concludes with a discussion of future work involving experimental exploration of ideas motivated by the neurophysiology of animal movement.

%\begin{itemize}
%\item {\em Depth discontinuities} associated with obstacle boundaries that may be difficult to distinguish from noise in optical flow,
%\item Moving objects in the field of view that produce localized optical flow that is inconsistent with the optical flow that is generated by self-motion,
%\item Ephemeral persistence of features within the field of view (FoV),
%\item Flow indeterminacy due to very sparse optical sensor data in part of the FoV,
%\item Flow indeterminacy due to very dense optical feature data in which features cannot be reliably matched frame-to-frame,
%\item Rotational movement of the optical sensor relative to the features being observed.
%\end{itemize}
%Feature points on objects that are in motion and changes in lighting are also important confounding factors, but these will not be considered in the present paper. %deformable objects!

\medskip

\section{Implementing sense-act cycle} \label{sec:sense-act}

Our previous work on $\tau$-based navigation introduced an idealized model called {\em Eulerian optical flow}, which by analogy with the Eulerian description of fluid flow, assumes that feature images stream continuously across all photo receptors such that values of $\tau$ at each photo receptor are continuously available, \cite{Baillieul2020}.  A number of $\tau$-based steering laws in the Eulerian setting have been shown to produce robustly stable steering in a variety of settings.
In order to obtain reliable visual information in actual implementations using \textit{perceived time-to-transit} relative to features in the environment, a less idealized {\em Lagrangian Optical Flow} (LOF) model, which is defined as the pattern of apparent motion of image objects between two consecutive frames caused by movement, 
is used (\cite{Baillieul2020}).  
Since one of the goals of this paper is to develop a navigation strategy capable of running in real-time on a robotic platform, we have created software that implements our navigation laws using ROS/Gazebo together with several well-known %consider advantages and disadvantages of the different 
%feature detectors, descriptors and matchers and of the 
Optical Flow estimation techniques. 

\begin{figure*}[h]
\begin{center}
\includegraphics[scale=0.32]{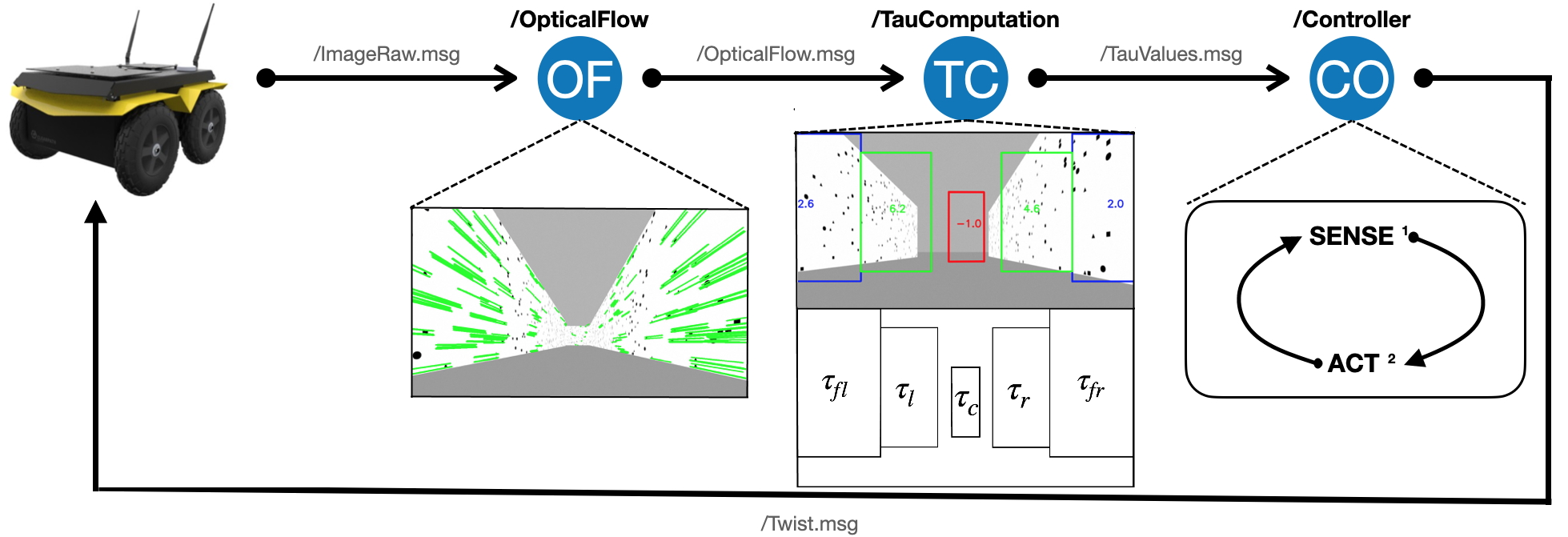}
\end{center}
\caption{Graphical representation of the ROS framework used in this paper.  Navigation is accomplished by message passing among nodes in the system ROS graph.}
\label{fig:bb:ROSarch}
\end{figure*}

%The Horn and Schunck method (\cite{hornschunk1981}) is susceptible to noise and it is computationally expensive. 
The Lucas-Kanade (LK) method (\cite{lucaskanade1981}) allows the computation of a sparse OF field (this means that OF vectors are estimated just for selected features in the image)  by optimizing an energy-like function at points in the image that have been determined by a feature detector. We choose this technique  because it is  fast  and  computationally  efficient. Since, by its nature, it works well for small movements but it fails when large pixel motion occurs, in our implementation it is used in its pyramidal form (\cite{pyramidalLK}).\
The features are detected by the Oriented FAST and Rotated BRIEF (ORB) detector because it is open source and suitable for real-time applications. The LK method allows to track features in subsequent frames but every time the number of remaining tracked features in the scene becomes lower than a threshold, the ORB detector is used again. 
%Exploiting the advantages of ROS, we develop a customized framework in order to achieve the final goal of autonomously guiding a mobile robot, equipped with a single monocular camera, in unknown environments.
In Fig. \ref{fig:bb:ROSarch} an outline of the ROS framework developed in this work is proposed with snapshots of the functionalities implemented in each node.

The Optical Flow node is responsible for the OF estimation. It acquires a sequence of images from the camera mounted on the robot and it extracts the relevant features to finally compute the optical flow vectors. 
The minimum quality and the maximum number of features to retain are parameters provided to the detector.
The node subscribes to the topic /camera\_used/image\_raw to get a camera frame and, thanks to the \textit{CVBridge} library, the ROS image is converted into a format manageable by \textit{OpenCV}. 

The goal of the {\em tau computation} node in Fig. \ref{fig:bb:ROSarch} is to analyze the array of keypoints with their velocities packed in the OF message, to compute $\tau$ values and to create the input signals for the controller. We transform the data coming from the Optical Flow node in order to make them coherent with a reference frame that has its origin at the intersection of the camera axis and the image plane. In this position, the origin of the coordinate frame is approximately at the Focus of Expansion (FOE), and since the precise estimation of the FOE is computationally expensive and sensitive to noise, and since our approximation has worked well in practice, we have used the camera-axis frame to determine $\tau$.

%If a forward-looking camera is going to be used to generate a steering signal based on balancing Optical Flow on left and right halves of the image, the most natural coordinate system is centered at the point where the optical axis intersects the image plane. 
Once all the data are expressed in the right reference frame, time-to-transit is computed as
\begin{equation*}
    |\tau_i| = \frac{\sqrt{x[i]^2 +y[i]^2}}{\sqrt{v_x[i]^2 + v_y[i]^2}}
\end{equation*}
where $x$, $y$, $v_x$ and $v_y$ are the arrays packed in the OF message. 
We define a fixed-size set of inputs that are given to the controller by analyzing the distribution of the tau values in the image space and of the optical flow field in different environments. We chose this approach to obtain with the goal of providing a good representation of the environment without over burdening the controller with computational effort.

To obtain more robust \textit{$\tau_{per}$} data we choose to divide the input image into five Regions of Interest (ROIs), as depicted in Fig. \ref{fig:bb:ROSarch}, which will output a single value each. The number and the dimension of these regions can be adapted to the specific environment in which we want to use the robot and the motion primitives that will be presented can be used regardless the number of regions.

Since the distribution of $\tau$ values is strongly related to the environment in which the robot is moving we can use it to select the right control law with the correct gains to be used.
When moving in a straight corridor the tau values are continuous in time and they generally come from every region except the central ROI (where time-to-transit is more sensitive to noise and may even be unavailable or unusably high).
The presence of a ninety degree turn, say at the end of a corridor, can be detected by a discontinuity in $\tau$ values, a suddenly unbalanced OF field, and small $\tau$ values in the central part of the FOV. 
There also exist  scenarios in which only the ROIs on one side of the environment provide enough features to compute reliable tau values.

\begin{figure}[h]
\begin{center}
\includegraphics[scale=0.18]{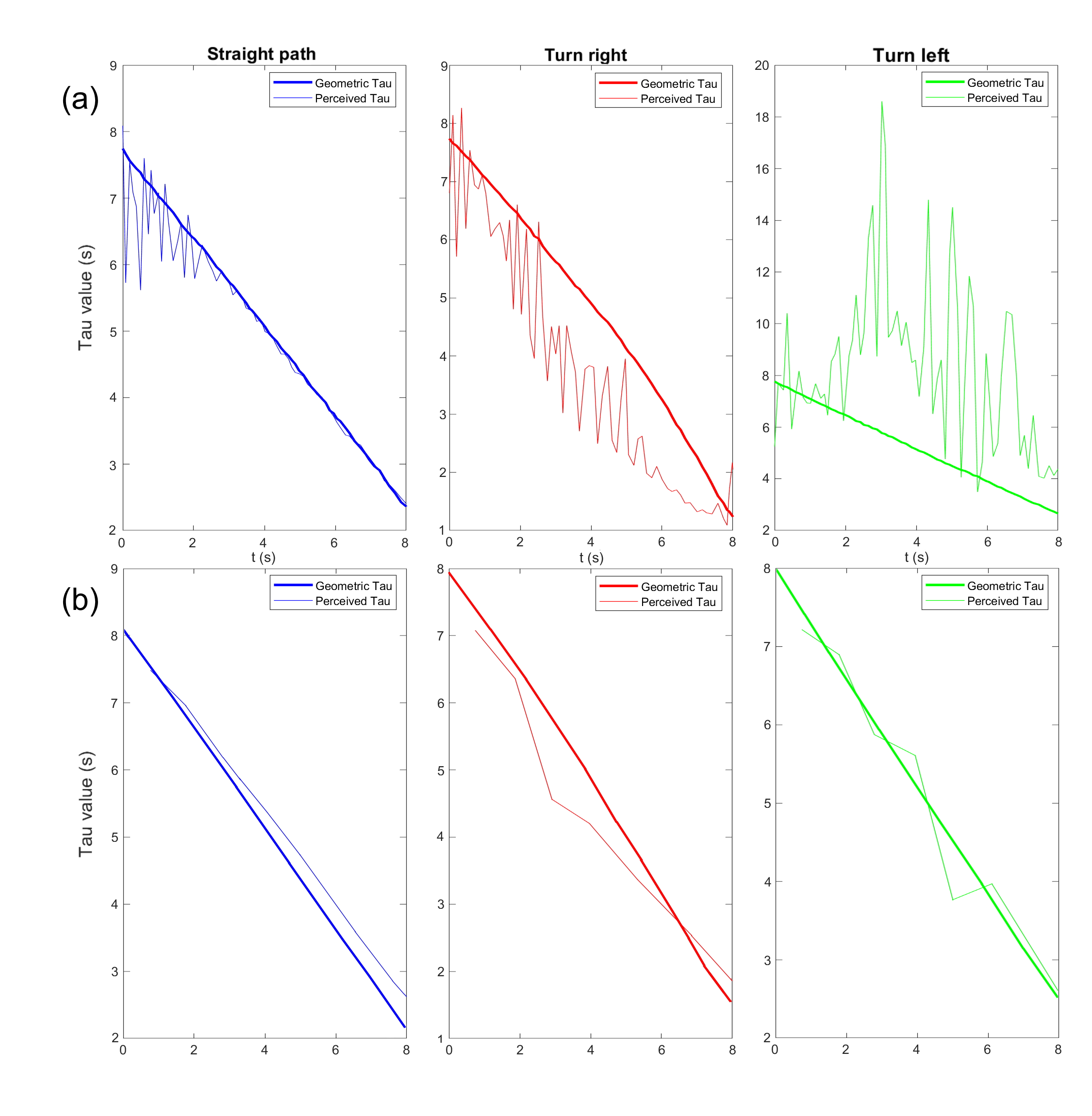}
\end{center}
\centering
\caption{Direct comparison between \textit{geometric} (thick line) and \textit{perceived} (fine line) \textit{time-to-transit} values during three tests, in (a) without and in (b) with the implementation of the sense-act interleaving.}
\label{fig:bb:NotSegmented}
\end{figure}

As noted in the introduction, camera rotation caused by a time-varying heading of the vehicle can cause large distortion of the estimate of $\tau$.
To make $\tau_{per}\approx\tau_{geom}$ a sense-act segmentation is implemented, which consists in segmenting the motion into alternating straight (sensing) and curved (acting) path segments.  Care must of course be taken to ensure that neither portion is sustained for too long a time interval to produce an instability.
An instantaneous sensing is possible only in an ideal scenario in which noise is not present. In a real environment the sense phase must be long enough to obtain a good estimation of $\tau$ values, but sensing for too much time means not being able to detect instantaneous variations of the OF field potentially losing important information coming from the environment. Moreover, the sense-act interleaving must guarantee the stability of the control law applied in the acting phase, and following a mathematical argument along the lines of \cite{Baillieul2020} (the arXiv version), it can be shown that if both the cycle length and act phase are of sufficiently short duration, the robot will stably and reliably move through the corridor---always tending to center itself.  This is checked with Matlab simulations assuming \textit{Eulerian} sensing and we can conclude that the stability is guaranteed for a wide large of sense-act duration values.
To understand how much this strategy can improve time-to-transit estimation a simple simulation in Gazebo using \textit{Lagrangian} sensing has been set up.
A feature is positioned in the left wall of a straight corridor and during three tests a robot collect information moving straight, then turning away from the feature and finally turning towards it. Fig. \ref{fig:bb:NotSegmented} shows a comparison between \textit{$\tau_{geo}$} and \textit{$\tau_{per}$}; in (a) no sense-act interleaving is used, while in (b) it is shown the significant improvement in perception of $\tau$ when a {\em sense-act} cycle is used (sense duration is $0.4s$, act duration is $0.25s$ and linear velocity $v=0.5m/s$).

\section{Steering control based on {\em time-to-transit}}\label{sec:motion-control}
When flying through a narrow passage, bees position themselves in the center of it in order to experience the same image velocity in both eyes. This characteristic of bee's flight has been highlighted by Srinivasan \textit{et al.} in \cite{srinivasan} and it represents the idea behind the \textit{Tau Balancing} control law introduced in \cite{sebestabaillieul} and further described in \cite{baillieul2019perceptual}. In this section we briefly describe how we adapt the motion primitives to the five ROIs. In particular, remembering (\ref{eq:jb:BasicVehicle}), in this work we choose to use: 
\begin{equation}
 \label{eq:tau_balancing_2}
 u(t)=k_f(\tau_{fl} - \tau_{fr}) + k_m(\tau_{l} - \tau_{r})
\end{equation}
where $\tau_{fl}$, $\tau_{fr}$, $\tau_r$ and $\tau_l$ are the time-to-transit value averages of the different ROIs of the image (the central ROI not included) as seen in the central panel of Fig. \ref{fig:bb:ROSarch}. It is possible to start from (\ref{eq:jb:BasicVehicle}) and isolate the following subsystem:
 \begin{equation}
    \begin{bmatrix}
    \dot x \\
    \dot \theta
    \end{bmatrix}
    =
    \begin{bmatrix}
    cos\theta \\
    k_f(\tau_{fl} - \tau_{fr})+k_m(\tau_l - \tau_r)
    \end{bmatrix}
    \label{eq:subsystem2}
\end{equation}

\begin{figure*}[h]
\begin{center}
\includegraphics[scale=0.32]{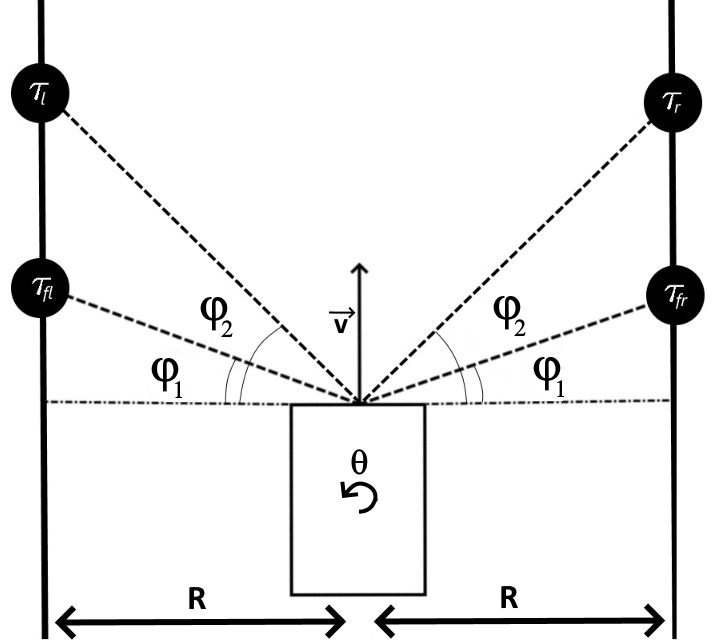}
\end{center}
\caption{Graphical representation of $\varphi_x$, $\theta$ and $R$.}
\label{fig:parameters1}
\end{figure*}

Adopting the Eulerian optical flow, a modified version of \textit{Theorem 1} in \cite{baillieul2019perceptual} can be written since in this case for any gain $k_{f}>0$ and $k_{m}>0$ there is an open neighborhood $U'$ of $(x, \theta)=(0, \frac{\pi}{2})$, $U'\subset \{(x, \theta) : -R < x < R; \varphi_2 < \theta < \pi - \varphi_2\}$ such that for all initial conditions $(x_0, y_0, \theta_0)$ with $(x_0, \theta_0) \in U'$, the \textit{Tau Balancing} in (\ref{eq:tau_balancing_2}) aligns the vehicle to the center of the corridor. The angle $\varphi_x$, $\theta$ and $R$ are graphically represented in Figure \ref{fig:parameters1}.

It is possible to linearize the subsystem in (\ref{eq:subsystem2}) about the rest point $(x, \theta)=(0, \frac{\pi}{2})$ obtaining:
\begin{equation*}
    \begin{bmatrix}
    \delta \dot x \\
    \delta \dot \theta
    \end{bmatrix}
    =
    \begin{bmatrix}
    0 & -1 \\
    2(f_fk_f + f_mk_m) & -2(k_ff_f^2+k_mf_m^2+k_fRf_f^2+k_mRf_m^2)
    \end{bmatrix}
    \begin{bmatrix}
    \delta x \\
    \delta \theta
    \end{bmatrix}
\end{equation*}
with $f_f=tan(\varphi_1)$ and $f_m=tan(\varphi_2)$. In this case it is like using two pinhole cameras with focal length $f_f$ and $f_m$. The eigenvalues of the matrix representing the controlled system are:
\begin{equation*}
    \lambda_{1,2}= -(f_f^2k_f+f_m^2k_f)(1+R) \pm \sqrt{(f_fk_f + f_mk_m)[(f_f^3k_f + f_m^3k_m)(1+R)^2-2]}
\end{equation*}
It can be noticed that they are always in the left half plane, proving the coefficient matrix is Hurvitz. Moreover, if the condition $(f_f^3k_f + f_m^3k_m)>\frac{2}{(1+R)^2}$ is met, the eigenvalues are real and negative meaning that the robot will not experience oscillations in aligning onto the center line.
Stability is guaranteed for every choice of $k_f$ and $k_m$, but these parameters must be tuned to obtain best performances. 

When features points are available only in the right or in the left part of the image, \textit{Tau Balancing} cannot be used.  
A possible solution had already been proposed by Kong \textit{et al.} in \cite{kong2013optical} observing the behavior of a species of bats, the \textit{Myotis Velifer}, in their natural habitat. The trajectories of these animals follow the edge of a wooded area thanks to the maximization of the difference in time-to-transit of features along the boundaries. 
This is the idea behind the formulation of the following motion primitive:
\begin{equation}
    u(t)=k[\tau_2^{\prime}(\theta(t))-\tau_1^{\prime}(\theta(t))]
    \label{eq:Kong_DiffMax}
\end{equation}
where $\tau_2$ and $\tau_1$ are the $time-to-transit$ values of two features on the same wall.
Unfortunately, since tau estimation is already in itself quite susceptible to noise, the computation of its derivative is generally quite poor and this makes it difficult to obtain the desired control action when the steering control law is used on a real platform. We introduce here a new control law called \textit{Single Wall} strategy which is able to cope with this problem that does not involve derivatives and it can then be written as:
\begin{equation}
 \label{eq:single_wall}
 u(t)=\pm k(\tau_x - c)
\end{equation}
where $\tau_x$ $\in\{\tau_{fl}, \tau_l, \tau_r, \tau_{fr}\}$ and $c$ is a constant (\cite{BorettiBich}). To demonstrate the stability of (\ref{eq:single_wall}), we analyze Eulerian models along the lines of \cite{Baillieul2020}, \cite{baillieul2019perceptual}.  It is possible to isolate from (\ref{eq:jb:BasicVehicle}) the following subsystem:
\begin{equation*}
\label{eq:sub_sw_left}
    \begin{bmatrix}
    \dot x \\
    \dot \theta
    \end{bmatrix}
    =
    \begin{bmatrix}
    cos\theta \\
    \pm k(\tau_x - c)
    \end{bmatrix}
\end{equation*}
and compute the rest point that, in this case, is $(x, \theta)=\left(\frac{\pm (c-fR-f)}{f}, \frac{\pi}{2}\right)$ depending on the chosen $\tau_x$. The constant $c$ is selected a priori depending on the general geometry of the environment, but a dynamic selection of $c$ is possible. Using Eulerian sensing, and linearizing the system around its rest point, we obtain:

\begin{equation*}
    \begin{bmatrix}
    \delta \dot x \\
    \delta \dot \theta
    \end{bmatrix}
    =
    \begin{bmatrix}
    0 & -1 \\
    fk & -kfc
    \end{bmatrix}
    \begin{bmatrix}
    \delta x \\
    \delta \theta
    \end{bmatrix}
\end{equation*}
The eigenvalues of the matrix representing the controlled system are:
\begin{equation}
\label{eq:eig_sw}
    \lambda_{1,2}= -\frac{kfc}{2} \pm \frac{\sqrt{kf(kfc^2-4)}}{2}
\end{equation}
Both of them are always in the left plane and for this reason the system is asymptotically stable. It is important to notice that if $k<\frac{4}{fc^2}$ the eigenvalues become complex numbers and this introduces oscillations in the system.

\section{Physics-based simulation and implementation on actual robots}

In this section we present some results from Gazebo simulations using Lagrangian optical flow models with relatively sparse sets of visual features and from the tests performed on a real robotic platform. 

%\subsection{Simulation Results}

\begin{figure}[h]
\includegraphics[scale=0.21]{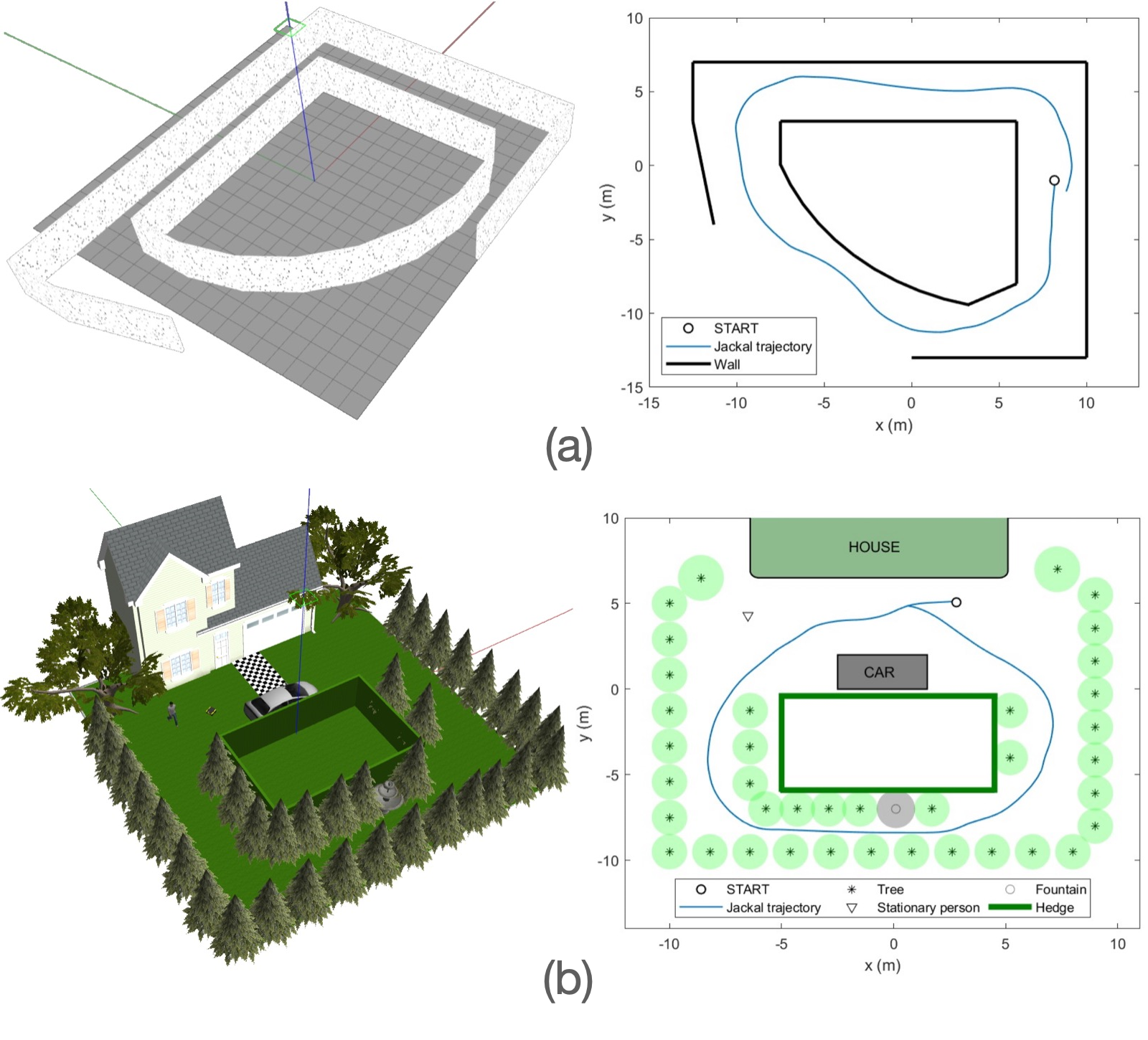}
\centering
\caption{(a) An artificial and (b) a realistic environment together with the trajectories followed by the robot during Gazebo simulations.}
\label{fig:bb:GazEnv}
\end{figure}
\textbf{A. \textit{Simulation Results:}} To verify the behaviour of the system presented in Fig. \ref{fig:bb:ROSarch} we run multiple Gazebo simulations in realistic and artificial environments and for each of them different geometric structures are considered (straight corridors, 90 degrees turns, corridors with multiple turns, single walls). Artificial environments have been used to verify the effectiveness of the software components in scenarios with a precise feature density. However, since the robot has to be able to navigate in the real world, in which it is not possible to define a fixed feature density, the realistic environments are used to test the real potential of the controlled system. Fig. \ref{fig:bb:GazEnv} provides an example of both types of environments together with the possible challenges that the robot has to overcome when it navigates using our control strategy. The trajectories followed by the robot are also reported and they show that the platform is able to safely navigate in these environments by switching among the control laws.

\begin{figure*}[h]
\begin{center}
\includegraphics[scale=.2]{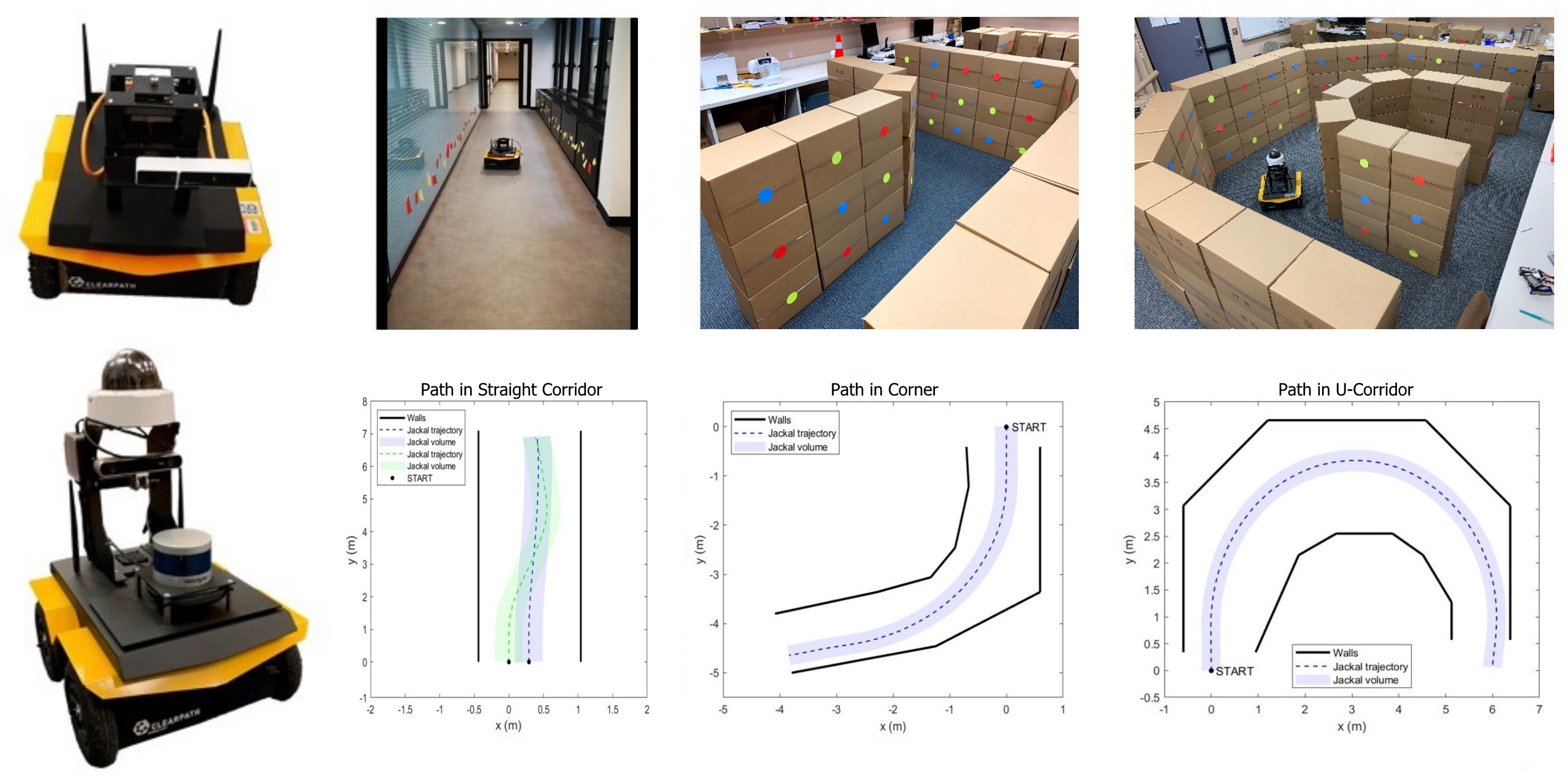}
\end{center}
\centering
\caption{On the left the Jackal UGV equipped with the MYNTEYE S1030 stereo camera (top image) and the Stereolabs ZED stereo camera (bottom image). On the right, some results from the tests run both at Boston University and at Politecnico di Torino.  A variety of environments were constructed using moving boxes.}
\label{fig:bb:realtest}
\end{figure*}

%\subsection{Experiments with a Jackal Robot UGV}

\textbf{B. \textit{Experiments with a Jackal Robot UGV:}} The promising results obtained with Gazebo simulations lead us to test our tau-based navigation strategy on real robots. The robotic platform we use is the Jackal UGV, an unmanned ground vehicle developed by Clearpath Robotics that includes an onboard computer, a GPS and an IMU fully integrated with ROS whose {skid-to-turn} steering is somewhat more challenging than the unicycle idealization (\ref{eq:jb:BasicVehicle}).  The extent of the difference between model (\ref{eq:jb:BasicVehicle}) and the actual robot has proven to be fairly negligible, and the control laws described above have worked well.

The Jackal robots have been equipped with stereo cameras (the MYNTEYE S1030 in Torino and the Stereolabs ZED 2 at BU).  We used the right monocular cameras as the only sensors in our navigation experiments. The most important characteristic is the horizontal field of view that has to be wide enough to recognize, with the proper timing, specific characteristics of the visual cues, e.g. discontinuities of the time-to-transit signal in environments containing turns. The two robots are shown in the left panel in Fig. \ref{fig:bb:realtest}.
We test the effectiveness of the entire algorithm by observing the behavior of the Jackal robot moving through different environments which have been set up at Politecnico di Torino and in the Boston University Robotics Lab. 
%For each scenario we collect the data referring to the position of the Jackal to plot the trajectories on Matlab. 
In each setting, we combine odometry and the IMU data to compute the position of the robot with respect to a coordinate frame that has its origin where the Jackal is powered on for the first time.
In the upper images of Fig. \ref{fig:bb:realtest} the different environment setups are shown, the features are mainly represented by the edges and the corners of the post-it notes but, since the tests have been conducted in real environments, there are also features belonging to other objects. The graphs in the lower panels of Fig.\ \ref{fig:bb:realtest} are based on odometry data that show that the {\em tau-balancing} control law, as predicted by the theory, steers the robot toward the center of the corridor and even successfully negotiates corners.

%This is a challenging scenario (variable feature density) and the robot must be able to navigate without registering any crash. The images in the lower part of Fig. \ref{fig:bb:realtest}  prove the effectiveness of the control strategy proposed in this work and demonstrate that it can be applied also in real scenarios since the robot is able to reach the center of the corridor, to maintain a certain distance (that depends on \textit{c} defined in eq. (\ref{eq:single_wall})) from a wall and to successfully perform corners.  
%{\color{blue} In light of what is above in Section III, I think a big contribution is what you have on ROI in your Ch. 5.}

%\medskip 

%{\color{blue} The control laws of Chs.\ 6 and 7 and the story about Eulerian models are good to include, but to some extent they need to be summarized and probably grouped together.}

\section{Conclusion and future work}

Within the page-count constraints imposed by the conference, we have reported selected results on reliable navigation of a mobile robot using the perceptual cue {\em time-to-transit}.  The work validates ideas with origins in the Cornell University Perception Lab of James J. Gibson, and at the same time it sets the stage for a broader investigation of vision based navigation using perceptual cues associated with optical flow.  In an expanded version of the paper, we shall report results on the use of a camera that is steered independently of the robot so as eliminate the confounding effects of movement along a curved path.  While a simple strategy of steering the camera so as to cancel the rotational component of movement can be shown to be effective, our current work is also aimed at understanding how camera movement might be used to support perceptual processes that involve numerous animal and human brain regions that are known to support spatial cognition, \cite{chen1994head},\cite{cho2001head},\cite{alexander2020egocentric}. In particular, our current research is being informed by investigations by Boston University colleagues (M.\ Betke and M.\ Hasselmo) who are working to correlate neural activity in running laboratory animals with 3D tracking of head and body poses recorded on thermal videos.  The results of this work will be reported soon.

%\newpage
%\onecolumn
%\begin{multicols}{2}    
\bibliography{references}

\begin{thebibliography}{99} 
%\bibitem{Leonard2008}
%J.~Leonard, J.~How, S.~Teller, M.~Berger, S.~Campbell, %G.~Fiore, L.~Fletcher, E.~Frazzoli, A.~Huang,
%S.~Karaman, {\em et al.} ``A perception-driven autonomous urban vehicle.'' in {\em J. of Field Robotics},
%25(10):727–774, 2008.
\bibitem{lee1976theory}
Lee, D.N., 1976. A theory of visual control of braking based on information about time-to-collision. Perception, 5(4), pp.437-459.https://doi.org/10.1068/p050437

\bibitem{lee1981plummeting}
Lee, D.N. and Reddish, P.E., 1981. ``Plummeting gannets: A paradigm of ecological optics.'' {\em Nature}, 293(5830), pp.293-294.
https://doi.org/10.1038/293293a0

\bibitem{baures2021time}
Baur\`es R, Fourteau M, Thébault S, Gazard C, Pasquio L, Meneghini G, Perrin J, Rosito M, Durand JB, Roux FE. Time-to-contact perception in the brain. J Neurosci Res. 2021 Feb;99(2):455-466. doi: 10.1002/jnr.24740. Epub 2020 Oct 18. PMID: 33070400.

\bibitem{tresilian1995perceptual}
Tresilian, J.R. Perceptual and cognitive processes in time-to-contact estimation: Analysis of prediction-motion and relative judgment tasks. Perception \& Psychophysics 57, 231–245 (1995). https://doi.org/10.3758/BF03206510

\bibitem{sebestabaillieul}
K. Sebesta and J. Baillieul. ``Animal-Inspired Agile Flight Using Optical Flow Sensing''. In: Proceeding of the 51$^{st}$ {IEEE} Conference on Decision and Control (CDC), 2012, pp. 3721-3727.

\bibitem{kong2013optical}
Z. Kong, K.\ {\"O}zcimder, N.W. Fuller, A. Greco, D. Theriault, Z. Wu, T. Kunz, M. Betke, and J. Baillieul. ``Optical flow sensing and the inverse perception problem for flying bats,'' In: Proceeding of the 52$^{nd}$ {IEEE} Conference on Decision and Control (CDC), 2013, pp. 1608-1615.

\bibitem{Baillieul2020}
J. Baillieul and F. Kang, ``Visual Navigation with a 2-pixel Camera---Possibilities and Limitations,''  In {\em Proceedings of the 21st IFAC World Congress} in Berlin, Germany, July 12-17, 2020. Also available from http://arxiv.org/abs/2103.00285.

\bibitem{lucaskanade1981}
B. D. Lucas and T. Kanade. ``An iterative image registration technique with an application to stereo vision''. In: IJCAI'81: Proceedings of the 7th International Joint Conference on Artificial intelligence, 2 (1981), pp. 674–679.

\bibitem{pyramidalLK}
J.Y. Bouguet. "Pyramidal Implementation of the Lucas Kanade Feature Tracker Description of the algorithm". Intel Corporation Microprocessor Research Labs, (2000). Available from http://robots.stanford.edu/cs223b04/algo\_affine\_tracking.pdf.

\bibitem{srinivasan}
M. V. Srinivasan, R. J. D. Moore, S. Thurrowgood, D. Soccol and D. Bland. ``From biology to engineering : insect vision and applications to robotics''. In: Frontiers in Sensing: From Biology to Engineering, Springer, 2012.

\bibitem{baillieul2019perceptual}
J. Baillieul, 2019, December. ``Perceptual Control with Large Feature and Actuator Networks.'' In Proceedings of the 2019 {\em IEEE 58th Conference on Decision and Control (CDC)} (pp. 3819-3826), DOI: 10.1109/CDC40024.2019.9029615.

\bibitem{BorettiBich}
C. Boretti and P. Bich, ``Dictionary of motion primitives for vision-based navigation using Optical Flow'', M.S. Thesis, Politecnico di Torino and Boston University, 2021.  Available at http://www.baillieul.org/Robotics/ThesisChiaraPhilippe.pdf.

\bibitem{chen1994head}
L.L. Chen, L.H. Lin, C.A. Barnes, B.L. McNaughton.``Head-direction cells in the rat posterior cortex'' {\em Exp Brain Res}, 101, 24–34 (1994). https://doi.org/10.1007/BF00243213

\bibitem{cho2001head}
J. Cho, P.E. Sharp, ``Head direction, place, and movement correlates for cells in the rat retrosplenial cortex,'' {\em Behav. Neurosci}. 115, 3–25 (2001).

\bibitem{alexander2020egocentric}
A.S. Alexander, L.C. Carsten, J.R. Hinman, F. Raudiesg, W. Chapman, and M.E. Hasselmo,  2020.  "Egocentric boundary vector tuning of the retrosplenial cortex," {\em Science advances}. 2020 Feb 1;6(8):eaaz2322. DOI:10.1126/sciadv.aaz2322

%%\bibitem{hornschunk1981}
%%B. K. Horn and B. G. Schunck. ``Determining optical flow''. In: Technical Symposium East, International Society for Optics and Photonics, 181 (1981), pp. 319–331.
\end{thebibliography}
\bibliographystyle{IEEEtran}

%\end{multicols}

\end{document}